\documentstyle[fleqn,12pt]{article}
\newtheorem{theorem}{Theorem}
\newtheorem{corollary}{Corollary}

\newtheorem{proposition}{Proposition}

\newcommand{\blackslug}{\mbox{\hskip 1pt \vrule width 4pt height 8pt 
depth 1.5pt \hskip 1pt}}
\newcommand{\QED}{\quad\blackslug\lower 8.5pt\null\par\noindent}
\newcommand{\proof}{\par\penalty-1000\vskip .5 pt\noindent{\bf Proof\/: }}
\newcommand{\ru}{\rule[-0.4mm]{.1mm}{3mm}}
\newcommand{\nni}{\ru\hspace{-3.5pt}}

\newcommand{\pre}{\hspace{0.28em}}

\newcommand{\NIm}{\pre\nni\sim}
\newcommand{\NI}{\mbox{$\: \nni\sim$}}
\newcommand{\notNIm}{\pre\nni\not\sim}

\newcommand{\ga}{\mbox{$\alpha$}}
\newcommand{\gb}{\mbox{$\beta$}}

\newcommand{\cC}{\mbox{${\cal C}$}}

\newcommand{\Cn}{\mbox{${\cal C}n$}}

\newcommand{\eqdef}{\stackrel{\rm def}{=}}

\newcommand{\rev}[3]{\mbox{$#1$$#2$$#3$}}
\newcommand{\bKs}[1]{\mbox{${\bf K}$$\!{\bf *}$${\bf #1}$}}
\begin{document}
	
\bibliographystyle{plain}

\title{Belief revision and rational inference
\thanks{This work was partially supported 
by the Jean and Helene Alfassa fund for 
research in Artificial Intelligence}
}
\author{Michael Freund \\ D\'{e}partement de Math\'{e}matiques, \\
Universit\'{e} d'Orl\'{e}ans, 45067 Orl\'{e}ans, C\'{e}dex 2 (France) 
\\ e-mail: freund@centre.univ-orleans.fr \and
Daniel Lehmann \\ Institute of Computer Science, \\
Hebrew University, Jerusalem 91904 (Israel) \\
e-mail: lehmann@cs.huji.ac.il
}
\date{August 12, 1994}
\maketitle

\begin{abstract}
The (extended) AGM postulates for belief revision seem to deal with
the revision of a given theory $K$ by an arbitrary formula $\varphi$,
but not to constrain the revisions of two different theories
by the same formula. 
A new postulate is proposed and compared with other similar postulates that
have been proposed in the literature.
The AGM revisions that satisfy this new postulate stand in one-to-one 
correspondence with the rational, consistency-preserving relations.
This correspondence is described explicitly.
Two viewpoints on iterative revisions are distinguished and discussed.
\end{abstract}
\section{Introduction}
\label{sec:intro}
\subsection{Belief Revision}
\label{subsec:beliefrev}
Belief revision is the study of the way an agent revises or should revise
its beliefs when acquiring new information.
A popular framework for this study has been initiated by Alchour\'{o}n,
Makinson and G\"{a}rdenfors in ~\cite{AM:82,AM:85,AGM:85}, in which a set
of rationality postulates for theory revision were put forward.
It assumes beliefs are sets of formulas closed under logical consequence
(i.e., theories) and that new information is a formula. The present work fits 
squarely in this framework.
An up-to-date description of this very active research area may be found 
in~\cite{Gard:beliefrev}. 
The basic assumptions underlying AGM's viewpoint may be summarized as follows.
\begin{itemize}
\item The agent holds beliefs: those beliefs constitute some logical theory.
There is no additional structure to the agent's theory. No beliefs are
stronger than others.
\item The agent's knowledge is changing but the world is not.
Keller and Winslett~\cite{KelWin:85} noticed that {\em updating}
an agent's knowledge about a changing world is a process that is altogether
different from theory revision.
Katsuno and Mendelzon~\cite{KatMend:92} proposed different postulates for 
belief update.
\item When a new piece of information $\varphi$ is presented to an agent that 
holds a theory $K$ with which $\varphi$ is inconsistent, 
the agent will give precedence to the information $\varphi$, 
over the theory $K$. The theory $K$ is considered as 
less important or less reliable
than the new information $\varphi$ with which it conflicts:
the agent revises a weakly held theory with a piece of more reliable
information.
This policy is obviously not the right one in every situation,
but AGM assume the agent must have some good reason to prefer $\varphi$ 
to $K$. As noticed just above, this reason is {\em not} that $\varphi$ 
describes some change in a world the previous state of which was
described by $K$.
\item When revising a theory $K$ with some formula $\varphi$, the agent
will try keep as much as possible of its previously held beliefs: $K$.
In particular, if $\varphi$ is consistent with $K$, the agent will keep
all of $K$ in its new belief set. This assumption will be called the
{\em maximal retention} assumption.
\end{itemize}
AGM did not describe any specific method to revise theories that would be
the {\em right} way of revising under the assumptions above.
They thought that there is probably no unique best way of revising.
Instead, they proposed a set of rationality postulates
they claimed any reasonable method for belief revision should satisfy.
The original postulates were extended in~\cite{Gard:intro}.
This work accepts, under the assumptions above, the extended AGM postulates.
These postulates are meant to express formally the basic assumptions described
above.
These postulates will be presented now and their adequacy in expressing those
basic assumptions will be discussed in section~\ref{subsec:additional}.
\subsection{The AGM postulates}
\label{subsec:AGM}
The AGM postulates for belief revision are listed below: 
\Cn\ denotes logical consequence and, for an arbitrary theory $K$
(i.e., $K$ is a set of formulas closed under logical consequence), 
\rev{K}{\!*}{\varphi} denotes the result of revising $K$ by $\varphi$.
The inconsistent theory, i.e., the full language, is denoted by $K_{\perp}$.
It is a legal argument for~$*$.
This paper could have been developed in the more demanding framework,
where the first argument of a revision, $K$, must be a consistent theory:
the results are essentially the same.

For a justification of these postulates, in addition to the papers cited above,
the reader may also 
consider~\cite{Mak:giveup,GardMak:88,Mak:recovery,GarMak:92,Rott:92}.
Since the notion of a revision is considered by
philosophers as not being primitive, but derived from that of
{\em theory contraction}, the justification of the postulates below is
generally given by justifying corresponding postulates for contractions
and translating back and forth with the help of the Levi and Gardenf\"{o}rs
identities.
Not all specialists are in complete agreement concerning the philosophical
underpinnings of theory revision, the status of the Levi and Gardenf\"{o}rs
identities, or of the postulates below.
The present work does not attack those deep questions directly, but
presents technical results concerning the postulates that, in turn,
help understand what are the processes that the postulates may reasonably be 
thought of as modelling and what are those they do not model satisfactorily.
For any theory $K$ and formulas $\varphi$ and $\psi$:
\[
\bKs{1} \ \ \ \   \rev{K}{\!*}{\varphi} {\rm \ is \ a \ theory.}
\]
\[
\bKs{2} \ \ \ \  \varphi \in \rev{K}{\!*}{\varphi}.
\]
\[
\bKs{3} \ \ \ \   \rev{K}{\!*}{\varphi} \subseteq \Cn(K , \varphi).
\]
\[
\bKs{4} \ \ \ \ {\rm If \ } \neg\varphi \not \in K, 
{\rm \ then \ } \Cn(K , \varphi) \subseteq \rev{K}{\!*}{\varphi}.
\]
\[
\bKs{5} \ \ \ \ {\rm If \ } \rev{K}{\!*}{\varphi} {\rm \ is \ inconsistent, \  then \ } 
\varphi {\rm \ is \ a \ logical \ contradiction.}
\]
\[
\bKs{6} \ \ \ \ {\rm If \ } \models \varphi \leftrightarrow \psi , 
{\rm \ then \ } \rev{K}{\!*}{\varphi} = \rev{K}{\!*}{\psi}.
\]
\[
\bKs{7} \ \ \ \  \rev{K}{\!*}{(\varphi \wedge \psi)} 
	\subseteq \Cn( \rev{K}{\!*}{\varphi} , \psi).
\]
\[
\bKs{8} \ \ \ \ {\rm If \ } \neg\psi \not \in \rev{K}{\!*}{\varphi} , 
{\rm \ then \ } \Cn(\rev{K}{\!*}{\varphi} , \psi) \subseteq \rev{K}{\!*}{(\varphi \wedge \psi)}
\]
\noindent
The reader may notice that, in each of the postulates in which the revision
symbol ($*$) appears more than once (\bKs{6}, \bKs{7} and 
\bKs{8}),
the theory that appears on the left of the different occurrences of the 
revision symbol is the same.
We take this as an indication that those postulates do not say much on the
way the revised theory \mbox{$\rev{K}{\!*}{\varphi}$} depends on $K$.
We shall come back to this point in section~\ref{subsec:additional}.
We shall call a revision \rev{K}{\!*}{\varphi} a {\em mild} revision,
when \mbox{$\neg \varphi \not \in K$} and a {\em severe} revision,
if \mbox{$\neg \varphi \in K$}.

The AGM postulates have been found to be intimately related to
rational consequence relations (see~\cite{MakGar:89}).
The translation is precisely described by the following Theorem.
\begin{theorem}
\label{the:Kfixed}
If $*$ is a revision operation that satisfies the AGM postulates,
\bKs{1}--\bKs{8}, and $K$ is a theory,
the relation $\NIm^{K,*}$ defined by
\begin{equation}
\label{eq:K*}
\ga \NIm^{K,*} \gb {\rm \ iff \ }  \gb \in \rev{K}{\!*}{\ga}
\end{equation}
is rational and consistency-preserving.
\end{theorem}
\proof
We just remind the reader that consistency preservation is the following 
property:
if \mbox{$\ga \NIm {\bf false}$}, then $\ga$ is a logical contradiction.
The definition of all the other properties mentioned in this proof may 
be found in~\cite{LMAI:92}.
Suppose $K$ is a theory and $*$ satisfies the AGM postulates.
The consequence relation $\NIm^{K,*}$ satisfies Reflexivity by \bKs{2},
Left Logical Equivalence by \bKs{6}, Right Weakening and And 
by \bKs{1},
Conditionalization (i.e., (S)) by \bKs{7}, and Rational Monotonicity by
\bKs{8}.
It is consistency-preserving by \bKs{5}.
Any such relation is rational.
\QED
Notice that postulates \bKs{3} and \bKs{4} are not used in the proof
of Theorem~\ref{the:Kfixed}.
This point will be discussed after we have proved 
Theorem~\ref{the:representation}.

One may expect the converse to Theorem~\ref{the:Kfixed} to hold:
if \NI\ is an arbitrary, rational and consistency-preserving relation,
then, there is a theory $K$ and a revision $*$ satisfying
\bKs{1}--\bKs{8} such that \mbox{$\alpha$ \NI $\beta$} iff 
\mbox{$\beta \in \rev{K}{\!*}{\alpha}$}.
This result holds.
It is not difficult to present a direct proof at this stage.
An indirect proof will be preferred and presented in 
Theorem~\ref{the:converse}. 
\subsection{Critique of the AGM postulates}
\label{subsec:additional}
The AGM postulates have been discussed in the literature, in particular
in~\cite{KatMend:91}, \cite{KatMend:92} and~\cite{DarwPearl:TARK}.
Some authors reject some of the postulates, in particular \bKs{8},
the same and others propose to extend the AGM postulates by additional 
postulates.
The main line of critique of the AGM postulates seems to be that they
do not constrain the revisions enough: some revisions allowed by the
postulates do not seem reasonable.
There are three points, in particular, on which it seems that the
postulates do not say enough, or, more precisely, one could feel that
additional acceptable postulates may be proposed.
\begin{itemize}
\item The postulates do not seem to enforce the principle of maximal retention
when the formula $\varphi$ is inconsistent with the theory $K$.
\item They do not seem to impose enough constraints on
the revision of two different theories by the same formula, or by related
formulas.
\item They do not seem to enforce enough constraints on iterated revisions, 
e.g., concerning the relation between $K$, $\varphi$, $\psi$ and 
\rev{(\rev{K}{\!*}{\varphi})}{*}{\psi}.
\end{itemize}
The proposals to drop some of the AGM postulates seem to stem from
the desire to add some other postulates with which they are inconsistent,
more than from a direct critique of the postulates in question.
The first point raised above has been the reason for the interest in
maxichoice contractions manifested by AGM, but no suitable additional
postulate has been proposed to deal with it.
The second point has been very aptly discussed in~\cite{MakGar:89}.
The authors write: ``Revision is an operation of two
arguments, forming $A*x$ out of theory $A$ and proposition $x$.
On the other hand, nonmonotonic inference conceived as an operation
$\cC(x)$ defined as \mbox{$\{ y \mid x \NIm y \}$} 
is a function of only one argument $x$. For this reason the logic of theory
change is potentially more general than the logic of nonmonotonic inference,
in that it allows the possibility of {\em variation} in the other
argument $A$. {\em Potentially}, because this possibility has hardly been 
explored. The postulates for revision presented in~\cite{Gardenfors:Flux} all
concern the case where the theory $A$ is held {\em constant}''.
We shall show, in the sequel, that the third point is intimately related
to the second one.
This paper will present an additional postulate constraining the way
one may revise different theories by the same formula.
Its main result is that this additional postulate precisely reduces the
generality of revisions in a way that makes revision isomorphic to
nonmonotonic inference. 
It will be shown that some of the additional postulates proposed
in the literature are incompatible with the AGM axioms and that those that
are not, are implied by our postulate.
\subsection{Plan of this paper}
\label{subsec:plan}
In section~\ref{sec:previous}, the additional postulates that were
previously proposed in the literature, by Katsuno and Mendelzon on one
hand, and by Darwiche and Pearl on the other hand, are described and 
discussed.
In section~\ref{sec:minimalin} the {\em minimal influence} postulate, 
\bKs{9}, is presented and justified on intuitive grounds. 
Some first consequences
of \bKs{9} that concern iterated revisions are then proven.
In section~\ref{subsec:K9U8} and \ref{subsec:K9DP}, its relations with the 
postulates of sections~\ref{subsec:KM} and~\ref{subsec:DP}, respectively,
are analyzed.
In section~\ref{subsec:representation}, 
we prove the main result of this paper:
revisions that satisfy \bKs{1} to \bKs{9} stand in one-to-one
correspondence with rational, consistency-preserving relations.
In section~\ref{subsec:conservext} a conservative extension result, 
a model-theoretic description of revision, and 
the converse to Theorem~\ref{the:Kfixed} are proven.
We discuss, then, in section~\ref{subsec:revwith}, the meaning of this 
result for the ontology of theory revision.
Section~\ref{sec:conc} discusses the existence
of two conflicting views on iterated revisions, concludes and describes some 
open questions.
\section{Additional postulates previously proposed}
\label{sec:previous}
\subsection{The Katsuno and Mendelzon postulate}
\label{subsec:KM}
In~\cite{KatMend:92}, Katsuno and Mendelzon considered the following postulate.
\[
{\bf U8} \ \ \ \rev{(K \cap K')}{*}{\varphi} = 
	(\rev{K}{\!*}{\varphi}) \cap (\rev{K'}{\!*}{\varphi}).
\]
\noindent
Reasonable as it seems, this postulate is nevertheless
inconsistent with the AGM postulates. 
This is probably the reason why Katsuno and Mendelzon dropped or weakened 
some of the AGM postulates in the final version of their paper.
The postulate that will be proposed in this paper is closely related to
${\bf U8}$, therefore we shall analyze it in some detail.
Our first remark is that any revision that satisfies the AGM postulates
also satisfies a special case (the {\em mild} case) of ${\bf U8}$.
\begin{proposition}
\label{prop:KM1}
If $*$ satisfies \bKs{3} and \bKs{4}, then, 
for any $\varphi$ such that \mbox{$\neg \varphi \not \in K \cup K'$},
\mbox{$\rev{(K \cap K')}{*}{\varphi} = 
(\rev{K}{\!*}{\varphi}) \cap (\rev{K'}{\!*}{\varphi})$}.
\end{proposition}
The proof is obvious.
Our second remark is that ${\bf U8}$ may be broken into two halves.
\begin{proposition}
\label{prop:KM2}
The postulate ${\bf U8}$ is equivalent to the following two properties:
\[
{\bf U8.1} \ \ {\rm If \ } K \subseteq K' , {\rm \ then \ } 
\rev{K}{\!*}{\varphi} \subseteq \rev{K'}{\!*}{\varphi} , {\rm \ and}
\]
\[
{\bf U8.2} \ \ (\rev{K}{\!*}{\varphi}) \cap (\rev{K'}{\!*}{\varphi}) \subseteq 
\rev{(K \cap K')}{*}{\varphi}.
\]
\end{proposition}
The proof is obvious.
Property ${\bf U8.1}$ has been named the postulate of Addition Monotonicity
and considered by G\"{a}rdenfors, Makinson and Segerberg 
in~\cite{Gard:Ramsey,Segerberg:impossibility,Mak:imposs}.
Despite its intuitive appeal (shouldn't the revised theory depend
monotonically on the revised theory?), it
has been shown to be inconsistent with the AGM postulates.
\begin{proposition}
\label{prop:KM3}
There is no revision operation that satisfies \bKs{4}, \bKs{5}
and ${\bf U8.1}$.
\end{proposition}
\proof
Suppose $*$ satisfies \bKs{4} and ${\bf U8.1}$.
Let $\varphi$ be an arbitrary formula that is not a logical contradiction.
By ${\bf U8.1}$, we have:
\[
\rev{\Cn(\varphi)}{*}{\bf true} \subseteq \rev{K_{\perp}}{\!*}{\bf true}.
\]
\noindent
But, since $\varphi$ is not a logical contradiction, 
\mbox{$\neg {\bf true} \not \in \Cn(\varphi)$} and, by \bKs{4},
we have
\mbox{$\Cn(\Cn(\varphi) , {\bf true}) \subseteq 
\rev{\Cn(\varphi)}{\!*}{\bf true}$}.
We conclude that 
\mbox{$\varphi \in \rev{\Cn(\varphi)}{\!*}{\bf true}$}, and, therefore,
\mbox{$\varphi \in \rev{K_{\perp}}{\!*}{\bf true}$}.
But the set of all formulas that are not a logical contradiction is
inconsistent and \rev{K_{\perp}}{\!*}{\bf true} is inconsistent,
contradicting \bKs{5}.
\QED
It will be shown in section~\ref{subsec:K9U8} that the additional
postulate we propose, \bKs{9}, implies ${\bf U8.2}$ 
and a special case of ${\bf U8.1}$.
\subsection{The Darwiche and Pearl postulates}
\label{subsec:DP}
In~\cite{DarwPearl:TARK}, Darwiche and Pearl proposed four additional
postulates. Notice that \rev{\rev{K}{\!*}{\psi}}{*}{\varphi}
means \rev{(\rev{K}{\!*}{\psi})}{*}{\varphi}.
\[
{\rm (C1)} \ \ \ \ {\rm If \ } \varphi \models \psi , {\rm \ then \ }
\rev{\rev{K}{\!*}{\psi}}{*}{\varphi} = \rev{K}{\!*}{\varphi}.
\]
\[
{\rm (C2)} \ \ \ \ {\rm If \ } \varphi \models \neg \psi , {\rm \ then \ }
\rev{\rev{K}{\!*}{\psi}}{*}{\varphi} = \rev{K}{\!*}{\varphi}.
\]
\[
{\rm (C3)} \ \ \ \ {\rm If \ } \psi \in \rev{K}{\!*}{\varphi} , {\rm \ then \ }
\psi \in \rev{\rev{K}{\!*}{\psi}}{*}{\varphi}.
\]
\[
{\rm (C4)} \ \ \ \ {\rm If \ } \neg \psi \not \in \rev{K}{\!*}{\varphi} , 
	{\rm \ then \ }
\neg \psi \not \in \rev{\rev{K}{\!*}{\psi}}{*}{\varphi}.
\]
\noindent
It will be shown in section~\ref{subsec:K9DP} that the postulates (C1), (C3) 
and (C4) are implied by our additional postulate \bKs{9}.
The justifications given by Darwiche and Pearl for those postulates are
therefore indirect justifications for \bKs{9}.
The postulate (C2), contrary to the claims of Darwiche and Pearl,
is inconsistent with the AGM axioms.
\begin{proposition}
\label{prop:DP}
There is no revision that satisfies \bKs{1}--\bKs{4} and {\rm (C2)}.
\end{proposition}
\proof
Suppose $*$ satisfies \bKs{1}--\bKs{4} and (C2).
Let $\varphi$ be any tautology, say {\bf true} and $\psi$ be any logical
contradiction, say {\bf false}.
We have \mbox{${\bf true} \models \neg {\bf false}$}.
The postulate (C2) therefore implies that
\mbox{$\rev{\rev{K}{\!*}{\bf false}}{*}{\bf true} = \rev{K}{\!*}{\bf true}$}.
But, for any theory $K$, 
\mbox{$\rev{K}{\!*}{\bf false} = K_{\perp}$} by \bKs{1} and \bKs{2}.
Therefore, for any $K$,
\mbox{$\rev{K}{\!*}{\bf true} = \rev{K_{\perp}}{\!*}{\bf true}$}.
But, by \bKs{3} and \bKs{4}, for any {\em consistent} theory $K$,
\mbox{$\rev{K}{\!*}{\bf true} = K$}.
Since \rev{K_{\perp}}{\!*}{\bf true} does not depend on $K$, we conclude
that all consistent theories are equal. A contradiction.
\QED
\section{The minimal influence postulate}
\label{sec:minimalin}
\subsection{The postulate}
\label{subsec:postulate}
The postulate we propose to add to the AGM postulates is the following.
\[
\bKs{9}   \ \ \ {\rm For \ any \ theories \ } K , K' {\rm \ such \ that \ } 
	\neg \varphi \in K , \neg \varphi \in K' \ 
\rev{K}{\!*}{\varphi} = \rev{K'}{\!*}{\varphi}.
\]
\noindent
The postulate \bKs{9} is obviously equivalent to:
if \mbox{$\neg \varphi \in K$}, then 
\mbox{$\rev{K}{\!*}{\varphi} = \rev{K_{\perp}}{\!*}{\varphi}$}.
Its meaning is that, if $\neg \varphi$ is an element of $K$, 
i.e., the revision of $K$ by $\varphi$ is a severe revision, 
then the result of revising $K$ by $\varphi$ does not depend on $K$.
Any revision that satisfies \bKs{3}, \bKs{4} and \bKs{9}, satisfies:
\begin{equation}
\label{eq:new}
\rev{K}{\!*}{\varphi} = 
	\left \{ \begin{array}{l}
	\rev{K_{\perp}}{\!*}{\varphi} {\rm \ if \ } \neg \varphi \in K \\
	\Cn(K , \varphi) {\rm \ otherwise.}
	\end{array} \right.
\end{equation}
Such a revision is therefore determined by its restriction to $K_{\perp}$.
\bKs{9} characterizes those revision systems in which
the revision $\rev{K}{\!*}{\varphi}$ depends mainly on $\varphi$ and only 
minimally on $K$ (in the case of a {\em mild} revision).
Remark also that, for any revision $*$ that satisfies \bKs{3} and \bKs{4},
the new postulate \bKs{9} is equivalent to:
\mbox{$\rev{K}{\!*}{\varphi} = \rev{\Cn(K , \varphi)}{*}{\varphi}$}.

The postulate \bKs{9}, at first sight, does not seem to be what one is 
looking for. 
First, it seems in danger of being inconsistent with the
AGM postulates. This concern will be addressed in full in 
section~\ref{sec:representation}.
But, also, one probably feels that the revisions of $K$ should
depend on $K$ in a major way, even when one considers a severe revision.
The feeling that some postulate should be added is widespread
and this paper will present formal arguments as to why \bKs{9} is the right
postulate to add.
Those reasons will not completely dissipate a first negative reaction
to \bKs{9}. The source of this clash between formal arguments and 
intuitive reaction probably lies in 
the philosophical underpinnings of theory revision and the general
framework chosen by AGM to study theory revision.
This paper claims that, in the framework chosen by AGM, \bKs{9} is
the right postulate to be added.
The AGM framework does not seem to be the one in which to study
iterated revisions.

We shall now present direct justification for \bKs{9}.
The postulate \bKs{9} is a special case of Darwiche and Pearl's postulate
(C2) described in section~\ref{subsec:DP}, assuming \bKs{1} and \bKs{2}.
Take, there, $\psi$ to be a logical contradiction. 
Since \mbox{$\rev{K}{\!*}{\psi} = K_{\perp}$},
one obtains \bKs{9}.
Anybody convinced by their defense of (C2) will endorse
\bKs{9}.
Our defense of \bKs{9} will be presented now.
Notice that \bKs{9} is equivalent to the following two properties:
they will be justified separately.
\[
\bKs{9.1}  \ \ \ {\rm For \ any \ theory \ } K {\rm \ such \ that \ } 
	\neg \varphi \in K , \ 
\rev{K}{\!*}{\varphi} \subseteq \rev{K_{\perp}}{\!*}{\varphi}
\]
\[
\bKs{9.2}  \ \ \ {\rm For \ any \ theory \ } K {\rm \ such \ that \ } 
	\neg \varphi \in K , \ 
\rev{K_{\perp}}{\!*}{\varphi} \subseteq \rev{K}{\!*}{\varphi}
\]
The first, \bKs{9.1} is relatively easy to justify.
It is a special case of ${\bf U8.1}$, i.e., the postulate of 
Addition Monotonicity. 
This postulate has been favorably considered by
many: it is very natural to hope that the more is believed before a revision,
the more is believed after.
None of the many articles cited above that
discuss this postulate rejects it on the grounds its meaning seems unwanted,
and the only problem found with it has been that it is inconsistent
with the AGM postulates. 
The special case proposed here, \bKs{9.1}, does not fall prey to this
criticism, it is consistent with the AGM postulates, and should be adopted.

The second, \bKs{9.2}, is more difficult to justify.
A formally weaker postulate, \bKs{9.2'}, 
will be presented and justified in a direct way. 
It will then been shown that, in the presence of other AGM postulates, 
it implies \bKs{9.2}.
The consideration of a similar property was suggested by Isaac Levi.
\[
\bKs{9.2'} \ \ \ \ 
	{\rm If \ } \psi \in K {\rm \ and \ }
	\psi \in \rev{K_{\perp}}{\!*}{\varphi} , {\rm \ then \ }
	\psi \in \rev{K}{\!*}{\varphi}.
\]
\noindent
Our justification for \bKs{9.2'} is the following.
If \mbox{$\psi \in K$}, then the principle of maximal retention implies
we should try to keep $\psi$ in \rev{K}{\!*}{\varphi}.
If \mbox{$\psi \in \rev{K_{\perp}}{\!*}{\varphi}$}, we know that
$\psi$ may be kept in \rev{K}{\!*}{\varphi} (in particular $\psi$ does 
not contradict $\varphi$),
\bKs{9.2'} says that, in this case, we should have 
$\psi$ in \rev{K}{\!*}{\varphi}.
It is easy to see that \bKs{9.2} implies \bKs{9.2'}, 
in the presence of \bKs{4}.
It will now be shown that \bKs{9.2'} implies \bKs{9.2}.
\begin{proposition}
\label{prop:9'9}
Any revision $*$ that satisfies \bKs{1}, \bKs{2} and \bKs{9.2'}
satisfies \bKs{9.2}.
\end{proposition}
\proof
Let $*$ satisfy \bKs{1}, \bKs{2} and \bKs{9.2'}.
Suppose \mbox{$\neg \varphi \in K$}, and 
\mbox{$\psi \in \rev{K_{\perp}}{\!*}{\varphi}$}.
We must show that
\mbox{$\psi \in \rev{K}{\!*}{\varphi}$}.
But, \mbox{$\varphi \rightarrow \psi \in \rev{K_{\perp}}{\!*}{\varphi}$}
and 
\mbox{$\varphi \rightarrow \psi \in K$}.
By \bKs{9.2'}, we conclude that
\mbox{$\varphi \rightarrow \psi \in \rev{K}{\!*}{\varphi}$}.
By \bKs{1} and \bKs{2}, then,
\mbox{$\psi \in \rev{L}{\!*}{\varphi}$}.
\QED

The rest of this paper is devoted to proving consequences of \bKs{9}:
the intuitive appeal of those consequences provides indirect justification
for it. In particular, in sections~\ref{subsec:K9U8} and~\ref{subsec:K9DP},
it will be shown that \bKs{9} implies most of the additional postulates 
that have been previously proposed in the literature.
All the arguments in favor of those postulates, in particular those developed
in~\cite{DarwPearl:TARK} for (C1), (C3) and (C4), provide indirect support
for \bKs{9}.
Our representation result, Theorem~\ref{the:representation}, and the analysis
of section~\ref{subsec:revwith} provide, both, a soundness result that
shows that many revisions satisfy \bKs{1}--\bKs{9}, and an ontology
for those revisions. This provides additional indirect justification
for \bKs{9}.

It seems that \bKs{9} contradicts, in a certain degree, the
assumption of maximal retention. If \mbox{$\neg \varphi \in K \cap K'$},
and \mbox{$\psi \in K$} but \mbox{$\psi \not \in K'$}, then
a revision $*$ that satisfies \bKs{9} will have to let go of
$\psi$ when revising $K$ by $\varphi$ or to let go of $\neg \psi$ when
revising $K'$ by $\varphi$.
The meaning of this remark is that one cannot retain maximally always,
one must compromise with the assumption of maximal retention in certain
cases to be able to apply it in other cases, as, e.g., in the justification
of \bKs{9.2'}.
\subsection{First consequences: iterated revisions}
\label{subsec:iterated}
Our understanding of the AGM framework, 
in which revisions are operations of two arguments,
a theory and a formula, does not require any special mention of iterated
revisions: the result of revising $K$ first by $\psi$ and then by $\varphi$
is the result of revising by $\varphi$ the theory that is the result of 
revising $K$ by $\psi$.
Some authors, for example~\cite{BouGold:AAAI93}, \cite{Bou:IJCAI93}
and~\cite{MaryAnneW:trans}, 
take a different view and prefer to treat revisions as operating on
a fixed theory and treat iterated revisions as a special case of varying
this theory. This attitude is methodologically at odds with the AGM point of 
view, and we shall argue, in section~\ref{subsec:viewpoints}, 
that it tries to answer a different question.
Discussion about this approach and its relations
to the present work is postponed to section~\ref{subsec:viewpoints}.

It is interesting to consider the meaning of our postulate
for iterated revisions.
There are two fundamental cases to consider: 
the case where revising by $\psi$
and then by $\varphi$ is equivalent to revising by the conjunction
$\psi \wedge \varphi$, and the case where it is equivalent to revising by
the second formula alone, $\varphi$.
Our result concerning the first case does not use \bKs{9}:
if the second revision is mild then the iterated revision is equivalent
to the direct revision by the conjunction.
\begin{proposition}
\label{prop:phiandpsi}
Let $*$ be a revision operation that satisfies \bKs{1}--\bKs{8}.
If \mbox{$\neg \varphi \not \in \rev{K}{\!*}{\psi}$}, then
\mbox{$\rev{(\rev{K}{\!*}{\psi})}{*}{\varphi} = 
\rev{K}{\!*}{(\psi \wedge \varphi)}$}.
\end{proposition}
\proof
Assume \mbox{$\neg \varphi \not \in \rev{K}{\!*}{\psi}$}.
By \bKs{7} and \bKs{8}, 
\mbox{$\Cn(\rev{K}{\!*}{\psi} , \varphi) = 
\rev{K}{\!*}{(\psi \wedge \varphi)}$}.
By \bKs{3} and \bKs{4},
\mbox{$\Cn(\rev{K}{\!*}{\psi} , \varphi) = 
\rev{(\rev{K}{\!*}{\psi})}{*}{\varphi}$}.
\QED
For the second case, we have three different results.
First, \bKs{9} alone implies that 
\mbox{$\rev{\rev{K}{\!*}{\psi}}{*}{\varphi} = \rev{K}{\!*}{\varphi}$}
if one has \mbox{$\neg \varphi \in \rev{K}{\!*}{\psi}$}
and \mbox{$\neg \varphi \in K$}.
Secondly, we have the following.
\begin{proposition}
\label{prop:psi}
Let $*$ be a revision operation that satisfies \bKs{1}--\bKs{9}.
If \mbox{$\neg \varphi \in \rev{K}{\!*}{(\psi \vee \varphi)}$}, then
\mbox{$\rev{(\rev{K}{\!*}{\psi})}{*}{\varphi} = \rev{K}{\!*}{\varphi}$}.
\end{proposition}
\proof
Suppose \mbox{$\neg \varphi \in \rev{K}{\!*}{(\psi \vee \varphi)}$}.
We are going to show that $\neg \varphi$ is an element of both
\rev{K}{\!*}{\psi} and $K$ and we shall conclude by \bKs{9}.
To show that \mbox{$\neg \varphi \in \rev{K}{\!*}{\psi}$}, we consider
two cases.
First, if \mbox{$\neg \psi \not \in \rev{K}{\!*}{(\psi \vee \varphi)}$},
by \bKs{8} and \bKs{6}, we have 
\[
\neg \varphi \in \Cn(\rev{K}{\!*}{(\psi \vee \varphi)} , \psi) \subseteq
\rev{K}{\!*}{((\psi \vee \varphi) \wedge \psi)}
= \rev{K}{\!*}{\psi}.
\]
\noindent
Secondly, if \mbox{$\neg \psi \in \rev{K}{\!*}{(\psi \vee \varphi)}$},
since $\neg \varphi$ and $\psi \vee \varphi$ are also elements of
\rev{K}{\!*}{(\psi \vee \varphi)} (by \bKs{2}), we conclude that
\rev{K}{\!*}{(\psi \vee \varphi)} is inconsistent.
By \bKs5, $\psi \vee \varphi$ is a logical contradiction and
$\neg \varphi$ is a tautology and, by \bKs1, an element of \rev{K}{\!*}{\psi}.
It is left to us to show that \mbox{$\neg \varphi \in K$}.
But, by \bKs{3}, 
\[
\neg \varphi \in \rev{K}{\!*}{(\psi \vee \varphi)} \subseteq 
\Cn(K , \psi \vee \varphi) \subseteq \Cn(K , \varphi).
\]
\noindent
We conclude that
\mbox{$\neg \varphi \in K$}.
\QED
The third result of this family is perhaps the most striking.
It generalizes the postulates (C1) and (C3) of Darwiche and Pearl.
\begin{proposition}
\label{prop:gen}
Let $*$ be a revision operation that satisfies \bKs{1}--\bKs{9}.
If \mbox{$\psi \in \rev{K}{\!*}{\varphi}$}, then
\mbox{$\rev{(\rev{K}{\!*}{\psi})}{*}{\varphi} = \rev{K}{\!*}{\varphi}$}.
\end{proposition}
The meaning of Proposition~\ref{prop:gen} is very natural: if revising $K$ 
by $\varphi$ would convince the agent that $\psi$ is true, then revising
$K$, first by $\psi$, and then by $\varphi$ amounts to revising, first
by $\psi$, and then by some information, $\varphi$, that conditionally 
implies the first information. This boils down to revising first by some
partial information and then by the full information. 
This is indeed expected to
be equivalent to revising directly by the full information.
\proof
Suppose \mbox{$\psi \in \rev{K}{\!*}{\varphi}$}.
Two cases will be considered.
Suppose, first, that \mbox{$\neg \varphi \not \in \rev{K}{\!*}{\psi}$}.
By Proposition~\ref{prop:phiandpsi}, 
\mbox{$\rev{(\rev{K}{\!*}{\psi})}{*}{\varphi} = 
\rev{K}{\!*}{(\psi \wedge \varphi)}$}.
But \mbox{$\psi \in \rev{K}{\!*}{\varphi}$}, and, by Theorem~\ref{the:Kfixed}
and the properties of rational relations,  
\mbox{$\varphi \NIm^{K , *} \psi$} implies that, for any $\chi$,
\mbox{$\varphi \NIm^{K , *} \chi$} iff 
\mbox{$\varphi \wedge \psi \NIm^{K , *} \chi$}.
Therefore 
\mbox{$\rev{K}{\!*}{\varphi} = \rev{K}{\!*}{(\varphi \wedge \psi)}$}.
We conclude that 
\mbox{$\rev{(\rev{K}{\!*}{\psi})}{*}{\varphi} = \rev{K}{\!*}{\varphi}$}.

Suppose, now, that \mbox{$\neg \varphi \in \rev{K}{\!*}{\psi}$}.
By \bKs{3}, \mbox{$\neg \varphi \in \Cn(K , \psi)$}
and \mbox{$\psi \rightarrow \neg \varphi \in K$}.
But \mbox{$\psi \in \rev{K}{\!*}{\varphi}$}, and, similarly,
\mbox{$\varphi \rightarrow \psi \in K$}.
We conclude that \mbox{$\neg \varphi \in K$}.
By \bKs{9}, then, 
\mbox{$\rev{(\rev{K}{\!*}{\psi})}{*}{\varphi} = \rev{K}{\!*}{\varphi}$}.
\QED
In the next sections, we shall examine in detail the relation between 
\bKs{9} and the postulates previously proposed by, both, 
Katsuno and Mendelzon, and Darwiche and Pearl.
\subsection{\bKs{9} and ${\bf U8}$}
\label{subsec:K9U8}
Our first result is that any revision that satisfies \bKs{9} satisfies
a special case of ${\bf U8}$.
\begin{proposition}
\label{prop:K9U8.1}
Let $*$ be any revision operation that satisfies \bKs{9}.
If \mbox{$\neg \varphi \in K \cap K'$}, then 
\mbox{$(\rev{K}{\!*}{\varphi}) \cap (\rev{K'}{\!*}{\varphi}) = 
\rev{(K \cap K')}{*}{\varphi}$}.
\end{proposition}
The proof is obvious. 
Our second result deals with ${\bf U8.2}$.
\begin{proposition}
\label{prop:K9U8.2}
Any revision operation $*$ that satisfies \bKs{3}, \bKs{4} 
and \bKs{9} satisfies ${\bf U8.2}$.
\end{proposition}
\proof
For any revision $*$ that satisfies the assumptions, we must show that
\mbox{$(\rev{K}{\!*}{\varphi}) \cap (\rev{K'}{\!*}{\varphi}) \subseteq 
\rev{(K \cap K')}{*}{\varphi}$}.
Suppose, first, that \mbox{$\neg \varphi \not \in K \cap K'$}.
By \bKs{4}, 
\mbox{$\Cn(K \cap K' , \varphi) \subseteq \rev{(K \cap K')}{*}{\varphi}$}.
By \bKs{3},
\[
(\rev{K}{\!*}{\varphi}) \cap (\rev{K'}{\!*}{\varphi}) 
\subseteq \Cn(K , \varphi) \cap 
\Cn(K' , \varphi) = \Cn(K \cap K' , \varphi).
\]
\noindent
Suppose, now, that \mbox{$\neg \varphi \in K \cap K'$}.
We conclude easily by Proposition~\ref{prop:K9U8.1}.
\QED
Our conclusion is that \bKs{9} implies all the cases of ${\bf U8}$
that are consistent with the AGM postulates,
the case that is inconsistent with the AGM postulates being, as shown 
in Proposition~\ref{prop:KM3}, the case in which $\neg \varphi$
is in $K$ but not in $K'$. 
In this case the revision of the intersection
may be larger than the intersection of the revisions.
\subsection{\bKs{9} and the Darwiche and Pearl postulates}
\label{subsec:K9DP}
Let us, now, consider the postulates proposed by Darwiche and Pearl.
We have seen, in Proposition~\ref{prop:DP}, that no revision may satisfy
(C2), but the following special case of (C2) is a consequence of
\bKs{9}.
\begin{proposition}
\label{prop:specC2}
Any revision operation $*$ that satisfies \bKs{1}, \bKs{2} and \bKs{9}
satisfies:
\[
{\rm (C2') \ \ \ \  If\ } \neg \varphi \in K {\rm \ and \ }
\varphi \models \neg \psi , {\rm \ then \ }
\rev{\rev{K}{\!*}{\psi}}{*}{\varphi} = \rev{K}{\!*}{\varphi}.
\]
\end{proposition}
\proof
Suppose \mbox{$\neg \varphi \in K$} and \mbox{$\varphi \models \neg \psi$}.
By \bKs{9}, it is enough to prove that 
\mbox{$\neg \varphi \in \rev{K}{\!*}{\psi}$}, but this is the case since
\mbox{$\psi \models \neg \varphi$}.
\QED
We shall now show that the other postulates considered by Darwiche and Pearl
are consequences of \bKs{9}.
\begin{proposition}
\label{prop:K9C}
Any revision operation $*$ that satisfies \bKs{1}--\bKs{9} satisfies
{\rm (C1)}, {\rm (C3)} and {\rm (C4)}.
\end{proposition}
\proof
For (C1), suppose \mbox{$\varphi \models \psi$}.
By \bKs{1} and \bKs{2}, \mbox{$\varphi \in \rev{K}{\!*}{\varphi}$}
and, by Propostion~\ref{prop:gen}, we conclude that
\mbox{$\rev{\rev{K}{\!*}{\psi}}{*}{\varphi} = \rev{K}{\!*}{\varphi}$}.

For (C3), assume that \mbox{$\psi \in \rev{K}{\!*}{\varphi}$}.
By Propostion~\ref{prop:gen},
\mbox{$\rev{\rev{K}{\!*}{\psi}}{*}{\varphi} = \rev{K}{\!*}{\varphi}$}.
We conclude that \mbox{$\psi \in \rev{\rev{K}{\!*}{\psi}}{*}{\varphi}$}.

For (C4), assume that \mbox{$\neg \psi \not \in \rev{K}{\!*}{\varphi}$}.
We must show that 
\mbox{$\neg \psi \not \in \rev{\rev{K}{\!*}{\psi}}{*}{\varphi}$}.
Assume, first, that \mbox{$\neg \varphi \not \in \rev{K}{\!*}{\psi}$}.
In this case, by Proposition~\ref{prop:phiandpsi}, 
\mbox{$\rev{\rev{K}{\!*}{\psi}}{*}{\varphi} = 
\rev{K}{\!*}{(\varphi \wedge \psi)}$}.
But \mbox{$\neg \psi \not \in \rev{K}{\!*}{\varphi}$}
and therefore, by \bKs{1}, \mbox{$ \varphi \not \models \neg \psi$}
and $\varphi \wedge \psi$ is not a logical contradiction.
By \bKs{5}, then, \mbox{$\rev{K}{\!*}{(\varphi \wedge \psi)}$} is consistent
and \mbox{$\neg \psi \not \in \rev{K}{\!*}{(\varphi \wedge \psi)}$}.
We conclude that 
\mbox{$\neg \psi \not \in \rev{\rev{K}{\!*}{\psi}}{*}{\varphi}$}.
Assume, then, that \mbox{$\neg \varphi \in \rev{K}{\!*}{\psi}$}.
By, \bKs{3}, \mbox{$\psi \rightarrow \neg \varphi \in K$}.
We shall show that \mbox{$\neg \varphi \in K$} and conclude, by \bKs{9},
that \mbox{$\rev{\rev{K}{\!*}{\psi}}{*}{\varphi} = \rev{K}{\!*}{\varphi}$}.
To show that \mbox{$\neg \varphi \in K$}, it is enough, by \bKs{4}, 
to show that \mbox{$K \not \subseteq \rev{K}{\!*}{\varphi}$}.
We shall show that 
\mbox{$\psi \rightarrow \neg \varphi \not \in \rev{K}{\!*}{\varphi}$}, i.e.,
\mbox{$\varphi \rightarrow \neg \psi \not \in \rev{K}{\!*}{\varphi}$}.
Since \mbox{$\varphi \in \rev{K}{\!*}{\varphi}$}, it is enough to show
that \mbox{$\neg \psi \not \in \rev{K}{\!*}{\varphi}$}, which holds 
by hypothesis.
\QED
As a corollary, we realize that (C1), (C3) and (C4) follow from (C2').
\begin{corollary}
\label{co:C}
Any revision operation $*$ that satisfies \bKs{1}--\bKs{8} and (C2'),
satisfies (C1), (C3) and (C4).
\end{corollary}
\proof
We shall show that $*$ satisfies \bKs{9} and conclude by 
Proposition~\ref{prop:K9C}.
Suppose \mbox{$\neg \varphi \in K$}.
Since, \mbox{$\varphi \models \neg {\bf false}$}, by (C2'),
we have 
\mbox{$\rev{K}{\!*}{\varphi} = \rev{\rev{K}{\!*}{\bf false}}{*}{\varphi}$}.
But, by \bKs{1} and \bKs{2}, \mbox{$\rev{K}{\!*}{\bf false} = K_{\perp}$}.
\QED
We have brought additional indirect justification for \bKs{9}:
it implies all the postulates proposed by Darwiche an Pearl, except (C2),
that has been proved inconsistent. Some special case of (C2) is also
implied by \bKs{9}. 
\section{Representation of revisions by rational relations}
\label{sec:representation}
\subsection{Representation result}
\label{subsec:representation}
We shall now prove the main result of this paper:
revision operations that satisfy \bKs{1}--\bKs{9} may be represented
by rational, consistency-preserving relations.
There is, even, a bijection between the former and the latter.
This result is an improvement on the results of~\cite{MakGar:89}, 
that described a bijection between AGM revisions of a fixed theory
$L$ and rational, consistency-presrving relations \NI\ such that
\mbox{$L = \{ \ga \mid {\bf true} \NIm \ga \}$}.
This correspondence shows that the postulate \bKs{9} is consistent
with the AGM postulates: there are revisions that satisfy \bKs{1}--\bKs{9},
as many as rational, consistency-preserving relations.
\begin{theorem}
\label{the:representation}
There exists a bijection between the set of revisions that satisfy the
postulates \bKs{1}--\bKs{9} and the set of 
consistency-preserving, rational relations.
This map associates to every such revision $*$ the relation 
$\NIm_{*}$ defined by:
\begin{equation}
\label{eq:NIm*}
\varphi \NIm_{*} \psi {\rm \ iff \ } \psi \in \rev{K_{\perp}}{\!*}{\varphi}. 
\end{equation}
The inverse map associates to any rational, consistency-preserving 
relation \NI\ the revision $*_{\sim}$
defined by: 
\begin{equation}
\label{eq:*sim}
\rev{K}{\!*_{\sim}}{\varphi} = \left\{ \begin{array}{l}
	\{ \psi \mid \varphi \NIm \psi \} {\rm \ if \ } \neg \varphi \in K \\
	\Cn(K , \varphi) {\rm \  otherwise.}
			\end{array}
		\right.
\end{equation}
\end{theorem}
Note that the relation $\NIm_{*}$ defined in~(\ref{eq:NIm*}) is none other than
the relation $\NIm^{K_{\perp},*}$ defined in Theorem~\ref{the:Kfixed}.
Note also that the revision $*_{\sim}$ defined in~(\ref{eq:*sim}) from
the relation \NI\ has a very natural meaning.
It says that, in the case of a mild revision, revise as mandated
by \bKs{3} and \bKs{4}, and in the case of a severe revision,
disregard $K$ (since it is contradicted by the more solid information
$\varphi$) and replace it by $\varphi$ together with all the formulas that 
the agent thinks are
usually true when $\varphi$ is.
\proof
Suppose first that $*$ is a revision that satisfies \bKs{1} to 
\bKs{9}, and let
us show that the relation $\NIm_{*}$ defined in~(\ref{eq:NIm*}) 
is rational and consistency-preserving.
This follows from Theorem~\ref{the:Kfixed} applied to $K_{\perp}$ and $*$,
since $\NIm_{*}$ is $\NIm^{K_{\perp},*}$. Note that \bKs{3},
\bKs{4} and \bKs{9} are not used in this part of the proof.

Suppose now that \NI\ is a consistency-preserving, rational relation
and let us show that the revision $*_{\sim}$ defined in~(\ref{eq:*sim}) 
satisfies \bKs{1}--\bKs{9}.
The cases of \bKs{1} (by right weakening and ``and''),  
\bKs{2} (by reflexivity), \bKs{3} (if \mbox{$\neg \varphi \in K$},
\mbox{$\Cn(K , \varphi) = K_{\perp}$}), \bKs{4} (by definition),
\bKs{6} (by left logical equivalence), and \bKs{9} 
(by definition) are easily dealt with.
Let us show that $*_{\sim}$ satisfies \bKs{5}.
Suppose \mbox{$\rev{K}{\!*_{\sim}}{\varphi}$} is inconsistent.
Then, either \mbox{$\varphi \NIm {\bf false}$} or 
\mbox{$\neg \varphi \not \in K$}
and \mbox{$\Cn(K , \varphi)$} is inconsistent.
The latter is impossible, therefore \mbox{$\varphi \NIm {\bf false}$}
and we conclude by consistency-preservation.
For \bKs{7}, we must show that 
\mbox{$\rev{K}{\!*_{\sim}}{(\varphi \wedge \psi)} \subseteq \Cn(\rev{K}{\!*_{\sim}}{\varphi} , \psi)$}.
We shall consider two cases.
Suppose first that  \mbox{$ \neg \varphi \in K$}. 
Then \mbox{$\neg(\varphi \wedge \psi) \in K $} too.
Suppose \mbox{$\chi \in \rev{K}{\!*_{\sim}}{(\varphi \wedge \psi)}$}.
Then, \mbox{$\varphi \wedge \psi \NIm \chi$}.
By conditionalization (i.e., rule S), 
we have \mbox{$\varphi \NIm \psi \rightarrow \chi$}
and, since \mbox{$ \neg \varphi \in K$}, 
\mbox{$\psi \rightarrow \chi \in \rev{K}{\!*_{\sim}}{\varphi}$}.
Therefore, \mbox{$\chi \in \Cn(\rev{K}{\!*_{\sim}}{\varphi} , \psi)$}.
Suppose now  that \mbox{$ \neg \varphi \not \in K$}.
Then, \mbox{$\rev{K}{\!*_{\sim}}{\varphi} = \Cn(K , \varphi)$}, and
\mbox{$\Cn(\rev{K}{\!*_{\sim}}{\varphi} , \psi) = \Cn(K , \varphi \wedge \psi)$}.
But we have already shown that $*_{\sim}$ satisfies \bKs{3},
therefore 
\[
\rev{K}{\!*_{\sim}}{(\varphi \wedge \psi)} \subseteq 
\Cn(K , \varphi \wedge \psi).
\]
\noindent
Finally, for \bKs{8}, suppose
\mbox{$\neg \psi \not \in \rev{K}{\!*_{\sim}}{\varphi}$}.
We have to show that 
\[
\Cn(\rev{K}{\!*_{\sim}}{\varphi} , \psi) 
\subseteq  \rev{K}{\!*_{\sim}}{(\varphi \wedge \psi)},
\]
\noindent
or, equivalently that if \mbox{$\psi \rightarrow \chi \in \rev{K}{\!*_{\sim}}{\varphi}$},
then, \mbox{$\chi \in \rev{K}{\!*_{\sim}}{(\varphi \wedge \psi)}$}.
We consider again two cases.
If, first, 
\mbox{$\neg\varphi\in K$}, then
\mbox{$\varphi \NIm \psi \rightarrow \chi $}
and also \mbox{$\varphi \notNIm \neg \psi$},
therefore, by rational monotonicity,
\mbox{$\varphi \wedge \psi \NIm \psi \rightarrow \chi $}
and, by ``and'' and right weakening,
\mbox{$\varphi \wedge \psi \NIm \chi $}.
But \mbox{$\neg\varphi\in K$} and therefore 
\mbox{$\neg(\varphi \wedge \psi) \in K$} and we conclude that
\mbox{$\chi \in \rev{K}{\!*_{\sim}}{(\varphi \wedge \psi)}$}.
The second case we must consider is: 
\mbox{$\neg \varphi \not \in K$}.
Then \mbox{$\rev{K}{\!*_{\sim}}{\varphi} = \Cn(K , \varphi)$}
and, by hypothesis, $\neg \psi$ is not a member of this set.
This implies that 
\mbox{$\neg \varphi \vee \neg \psi \not \in K$}.
Therefore,
\mbox{$\rev{K}{\!*_{\sim}}{(\varphi \wedge \psi)} 
= \Cn(K , \varphi \wedge \psi)$}.
Therefore 
\mbox{$\rev{K}{\!*_{\sim}}{\varphi} = \Cn(K , \varphi) 
\subseteq \rev{K}{\!*_{\sim}}{(\varphi \wedge \psi)}$}.

To complete the proof of the representation theorem, we have to show that
the maps \mbox{$\NIm \longmapsto *_{\sim}$} and 
\mbox{$* \longmapsto \NIm_{*}$} are inverse maps, i.e., that
\mbox{$\NIm_{*_{\sim}} = \NIm$} and that 
\mbox{$*_{\sim_{*}} = *$}.
For the first equality, we have
\mbox{$\varphi \NIm_{*_{\sim}} \psi$} iff
\mbox{$\psi \in \rev{K_{\perp}}{\!*_{\sim}}{\varphi}$},
iff \mbox{$\varphi \NIm \psi$} since \mbox{$\neg \varphi \in K_{\perp}$}.
For the second equality, we have 
\[
\rev{K}{\!*_{\sim_{*}}}{\varphi} = \left\{ \begin{array}{l}
	\{ \psi \mid \varphi \NIm_{*} \psi \} {\rm \ if \ } 
			\neg \varphi \in K \\
	\Cn(K , \varphi) {\rm \  otherwise.}
			\end{array}
		\right.
\]
\noindent
Therefore,
\[
\rev{K}{\!*_{\sim_{*}}}{\varphi} = \left\{ \begin{array}{l}
	\rev{K_{\perp}}{\!*}{\varphi} {\rm \ if \ } 
			\neg \varphi \in K \\
	\Cn(K , \varphi) {\rm \  otherwise.}
			\end{array}
		\right.
\]
\noindent
But, if \mbox{$\neg \varphi \in K$}, by \bKs{9}, 
\mbox{$\rev{K_{\perp}}{\!*}{\varphi} = \rev{K}{\!*}{\varphi}$},
and if \mbox{$\neg \varphi \not \in K$},
by \bKs{3} and \bKs{4}, 
\mbox{$\Cn(K , \varphi) = \rev{K}{\!*}{\varphi}$}.
Notice that the postulates used now, \bKs{3}, \bKs{4} 
and \bKs{9} are those that were not used in the first part of this proof.
\QED
In section~\ref{subsec:AGM}, in the discussion following the proof of
Theorem~\ref{the:Kfixed}, we noticed that any theory $K$ and any revision
operation $*$ that satisfies \bKs{1}, \bKs{2} and \bKs{5}
to \bKs{8} defines a rational, consistency-preserving relation,
the relation $\NIm^{K , *}$ defined in formula~(\ref{eq:K*}).
We have shown, in Theorem~\ref{the:representation}, that any such relation 
is defined in this way by a specific theory $K_{\perp}$ and some
revision operation that satisfies not only \bKs{1}, \bKs{2} and 
\bKs{5}--\bKs{8}, but also \bKs{3}, \bKs{4} and \bKs{9}.
It does not mean that \bKs{3}, \bKs{4} and \bKs{9} are redundant, i.e.,
derivable from the other postulates, but only that the mapping
\mbox{$(K , *) \longmapsto \NIm^{K , *}$} is not injective.
In fact, the reader will quickly realize that no postulate is redundant, 
though \bKs{3} may be weakened to 
\mbox{$\rev{K}{\!*}{\bf true} \subseteq K$} 
in the presence of \bKs{6} and \bKs{7}.
We take this to mean that postulates \bKs{3}, \bKs{4} and \bKs{9}
say nothing about the {\em nonmonotonic inference} aspect of revision:
they are orthogonal to it.
\subsection{Consequences of the representation theorem}
\label{subsec:conservext}
The first consequence that we shall draw from Theorem~\ref{the:representation}
is that the postulate \bKs{9} does not constrain the ways one may revise 
a fixed theory $K$ more than the AGM postulates already did:
it only constrains the ways one may revise different theories by 
related formulas.
If we consider an AGM revision $*$ and a fixed theory $K$,
there is a revision that satisfies \bKs{1}--\bKs{9} and
that revises the theory $K$, for any formula $\varphi$, exactly as does $*$.
\begin{theorem}
\label{the:conservext}
Let $*$ a revision that satisfies the AGM postulates
\bKs{1} to \bKs{8}.
Then, for any theory $K$,
there exists a revision  $*_{K}$ that satisfies
\bKs{1}--\bKs{9} and such that,
for any $\varphi$, \mbox{$\rev{K}{\!*}{\varphi} = 
\rev{K}{\!*_{K}}{\varphi}$}.
\end {theorem}
\proof
Let $*$ a revision that satisfies the AGM postulates
\bKs{1} to \bKs{8} and $K$ an arbitrary theory.
By Theorem~\ref{the:Kfixed}, the relation $\NIm^{K , *}$, defined in
equation~(\ref{eq:K*}) is rational and consistency-preserving.
For simplifying notations, we shall denote $\NIm^{K , *}$ by \NI.
By Theorem~\ref{the:representation}, the revision $*_{\sim}$,
defined in equation~\ref{eq:NIm*},
satisfies \bKs{1}--\bKs{9}. 
But, for any $\varphi$,
\[
\rev{K}{\!*_{\sim}}{\varphi} = 
	\left \{ \begin{array}{l}
	\{ \psi \mid \varphi \NIm^{K , *} \psi \} 
		{\rm \ if \ } \neg \varphi \in K \\
	\Cn(K , \varphi) {\rm \  otherwise.}
			\end{array}
		\right.
\]
\noindent
Therefore, by equation~(\ref{eq:K*}),
\[
\rev{K}{\!*_{\sim}}{\varphi} = 
	\left \{ \begin{array}{l}
	\rev{K}{\!*}{\varphi} {\rm \ if \ } \neg \varphi \in K \\
	\Cn(K , \varphi) {\rm \  otherwise.}
			\end{array}
		\right.
\]
\noindent
But the right hand-side is equal to \rev{K}{\!*}{\varphi}.
\QED
From the proof above, one sees that the revision $*_{K}$ is defined by:
\begin{equation}
\label{eq:*L}
\rev{L}{\!*_{K}}{\varphi} = \left\{ \begin{array}{l}
	\rev{K}{\!*}{\varphi} {\rm \ if \ } \neg \varphi \in L \\ 
	\Cn(L , \varphi) {\rm \ otherwise.} 
			\end{array}
		\right.
\end{equation}
\noindent
An model-theoretic description of $*$ will be given now.
Suppose $*$ satisfies \bKs{1}--\bKs{9}, and $\NIm_{*}$ is the rational, 
consistency-preserving relation associated to it by equation~(\ref{eq:NIm*}).
Let $M$ be any ranked model that defines $\NIm_{*}$ (see~\cite{LMAI:92}).
Note, that, since $\NIm_{*}$ is consistency-preserving, every propositional
model is the label in $M$ of some state.
For any theory $K$, let us define $M_{K}$ as the ranked model obtained
from $M$ by creating, below the states of lowest rank in $M$, a new level
containing a state for each propositional model of $K$ (and labeled by this 
model).
Note that $M$ = $M_{K_{\perp}}$.
It is clear that $M_{K}$ is a ranked model.
Let $\NIm_{M_{K}}$ be the rational relation defined by $M_{K}$.
\begin{proposition}
\label{prop:modelth}
\[
\psi \in \rev{K}{\!*}{\varphi} {\rm \ iff \ } \varphi \NIm_{M_{K}} \psi
\]
\end{proposition}
\proof
The result follows from equation~(\ref{eq:new}) and the definition of
$\NIm_{M_{K}}$.
Note that there is some state at the lowest rank in $M_{K}$ that satisfies
$\varphi$ iff \mbox{$\neg \varphi \not \in K$}.
In this case \mbox{$\varphi \NIm_{M_{K}} \psi$} iff 
\mbox{$\psi \in \Cn(K , \varphi)$}.
\QED
An important remark, that will be developed in 
section~\ref{subsec:viewpoints},
is that, even though $M$ (and $\NIm_{*}$) may be recuperated back 
from $M_{K}$ by leaving out the states of lowest rank, 
there is not enough information in $\NIm_{M_{K}}$ to determine
$\NIm_{*}$, or $\NIm_{M_{L}}$ for \mbox{$L \neq K$}.
Think that one may be build a model $M'_{K}$ from $M_{K}$ by leaving out
all the states of $M_{K}$ that satisfy $K$ except the states of lowest rank.
The model $M'_{K}$ defines the same relation as $M_{K}$, but no
trace is left in it of the original ranks in $M$ of the states that satisfy
$K$.
Our next consequence of the representation theorem is the converse
of Theorem~\ref{the:Kfixed}.
\begin{theorem}
\label{the:converse}
Let \NI\ be an arbitrary, rational, consistency-preserving relation.
There is a theory $K$ and a revision $*$ that satisfies \bKs{1}--\bKs{9}
such that \mbox{$\varphi \NIm \psi$} iff 
\mbox{$\psi \in \rev{K}{\!*}{\varphi}$}.
\end{theorem}
\proof
Let \NI\ be as above.
Take \mbox{$K \eqdef \{ \alpha \mid {\bf true} \NIm \alpha \}$}.
It is clear that $K$ is a theory.
Take \mbox{$ * \eqdef *_{\sim}$}, defined in equation~\ref{eq:*sim}.
For any $\varphi$, we have:
\[
\rev{K}{\!*}{\varphi} = 
	\left \{ \begin{array}{l}
	\{ \psi \mid \varphi \NIm \psi \} {\rm \ if \ } \neg \varphi \in K \\
	\Cn(K , \varphi) {\rm \  otherwise.}
			\end{array}
		\right.
\]
\noindent
By the definition of $K$:
\[
\rev{K}{\!*}{\varphi} = 
	\left \{ \begin{array}{l}
	\{ \psi \mid \varphi \NIm \psi \} 
		{\rm \ if \ } {\bf true} \NIm \neg \varphi \\
	\Cn(K , \varphi) {\rm \  otherwise.}
			\end{array}
		\right.
\]
\noindent
But \NI\ is rational and: 
\mbox{$\rev{K}{\!*}{\varphi} = \{ \psi \mid \varphi \NIm \psi \} $}.
\QED
\subsection{Revising with a rational, consistency-preserving relation}
\label{subsec:revwith}
The definition of the revision $*_{\sim}$ in equation~(\ref{eq:*sim})
proposes a certain ontology for theory revision.
The agent has a certain rational, consistency-preserving relation
in mind. This relation describes, as far as the agent knows, 
what follows from what by default, or in normal circumstances.
When it receives some new information $\varphi$, the agent revises 
its current belief set $K$ by either adding $\varphi$ to its current belief
set, if $\varphi$ does not contradict $K$, or, if it contradicts $K$,
by forgetting about $K$ altogether and adopting $\varphi$ and all default
assumptions that go with $\varphi$ as its new belief set.
In the latter situation, the agent adopts not only $\varphi$ but all its
usual, normal consequences also.
This process seems very reasonable: when learning $\varphi$ that contradicts
its current beliefs $K$, the agent should probably not try to throw
away as little of possible of $K$ to accommodate just $\varphi$ but should
throw away enough of $K$
to accommodate also the most plausible consequences of $\varphi$.
Remember that it is a basic assumption of the AGM approach that
$\varphi$ is more reliable than $K$, so we are only saying that the
{\em default}, i.e. {\em normal}, consequences of $\varphi$ should
also be considered more secure than $K$.
If, for example, one's theory says it should rain in Paris and in Orl\'{e}ans
and one learns that there are no clouds in the sky of Paris, 
one has a choice between two revisions:
conclude that it does not rain in Paris but rains in Orl\'{e}ans,
or conclude that it rains neither in Paris nor in Orl\'{e}ans.
If one holds the default assumption that when the sky
is cloudless in Paris it does not rain in Orl\'{e}ans, 
the second way of revising seems more reasonable: it gives precedence to
the default ``when there are no clouds in Paris, it does not rain in 
Orl\'{e}ans'' to the proposition ``it rains in Orl\'{e}ans'' 
that was in the original theory.
Note that this does not contradict the principle of maximal retention,
since the original theory contained both 
``when there are no clouds in Paris, it does not rain in Orl\'{e}ans''
and ``it rains in Orl\'{e}ans'', and one could not keep both of them
in the face of the information that ``there are no clouds in Paris''.

In the process of iterated revisions considered here, the agent
revises its beliefs with new information, using its knowledge of
how things generally, or normally, behave. This knowledge is encoded
in a rational, consistency-preserving relation.
The agent does {\em not} revise its relation, i.e., its default
knowledge.
There is a lot to be said for this analysis: default assumptions
are typically much more stable than beliefs, and so it should be.
Consider, for example, the physicist revising his/her theory about
the world in light of the results of new experiments.
The method used in performing this revision, i.e., deciding what to
keep and what to throw away of the old theory, is a matter of meta-principles,
a methodological question, not a question of physics.
There is no reason to think that the result of this new experiment will
change anything in the methodological principles our physicist
is using.
Nevertheless, there must also be another process at work:
the revision (or the creation) of default assumptions.
After all, those default assumptions, or these meta-principles come
from the observation of evidence.
Decision theory seems to be mainly interested in this process of formation
or revision of default assumptions in the light of new evidence.
It operates on a longer time scale than the revisions considered here
and this paper has nothing to say about it.
The revisions we consider are revisions of belief sets performed under
a fixed set of default assumptions.
\section{Closing remarks}
\label{sec:conc}
\subsection{Two viewpoints on iterated revisions}
\label{subsec:viewpoints}
We have described and developed what we think is the original
AGM viewpoint concerning revisions. Revisions have two arguments:
a theory to be revised and a formula by which it is revised.
Iterated revisions do not deserve any special consideration.
Suppose an agent holds $*$ as its revision procedure.
If, at some point in time, it has belief set $K$ 
and learns $\varphi$ it will, then, acquire belief set \rev{K}{\!*}{\varphi}.
If the agent later learns $\psi$, it will just use the {\em same} revision $*$
to revise its current belief set \rev{K}{\!*}{\varphi}, by the new information
$\psi$, thus obtaining \rev{\rev{K}{\!*}{\varphi}}{*}{\psi} as its new
belief set. In this process, the belief revision process, $*$, 
{\em does not change}. From this point of view, the revision process
is fixed: {\em theories} are revised but the 
{\em revision method} stays fixed.
This is the {\em static} viewpoint.

There is a different point of view, that has been illustrated in a number
of recent works on iterated revisions, notably~\cite{BouGold:AAAI93}, 
\cite{Bou:IJCAI93} and~\cite{MaryAnneW:trans}.
This point of view seems to be rooted in Bayesian decision theory
and to draw on~\cite{Spohn:87}.
This alternative, {\em dynamic}, point of view has attracted most of 
the attention recently given to theory revision and can be described as
follows.
The agent starts with a belief set $K$ and a method for revising $K$
when some new information will come along, i.e., the agent knows only how to
revise $K$, its belief set, it does not know how to revise arbitrary theories.
If some new information comes along, it will revise $K$ with this new 
information $\varphi$, following its method and obtain a new belief set $K'$, 
but it will also adapt its revision method to be able to revise $K'$
when new information will eventually come along.
From this point of view, the new information $\varphi$ does not only
modify the belief set it {\em also modifies the revision method}.
This dynamic viewpoint presents a departure from the AGM static framework.
It even seems that those two points of view are incompatible.
It will be shown, first, that a dynamical viewpoint does not provide
a static revision operator, and then, that a static revision operator
does not provide a method for dynamically adapting revisions.

Let us take the dynamical viewpoint.
\begin{itemize}
\item In the dynamic view the new revision procedure
by which \rev{K}{\!*}{\varphi} will be revised depends on $\varphi$,
whereas, in the static view it should not.
\item It must not be the case that all theories may be obtained starting 
from $K$ by iterated revisions, whereas the static view insists on being
capable of revising arbitrary theories.
\item It must not be the case that the revision of a theory by a formula
is determined by the theory and the formula. It may depend on the
sequence of revisions that led to the theory.
In other terms, it may happen that $K$ is identical with
\rev{\rev{K}{\!*}{\varphi}}{*}{\psi} without implying that
\rev{K}{\!*}{\chi} is identical with 
\rev{\rev{\rev{K}{\!*}{\varphi}}{*}{\psi}}{\!*}{\chi}.
\end{itemize}

Let us take the static viewpoint, now.
The static viewpoint proposes \rev{\rev{K}{\!*}{\psi}}{*}{\varphi}
for the result of revising K, successively, first by $\psi$ and then
by $\varphi$. The dynamic viewpoint says this iterated revision should
be the result of revising \rev{K}{\!*}{\psi} by $\varphi$, {\em using the
revision method obtained by adapting the revision 
\mbox{$\mu \eqdef \chi \longmapsto \rev{K}{\!*}{\chi}$} to the new information
$\psi$}.
But, in section~\ref{subsec:conservext}, 
following Proposition~(\ref{prop:modelth}), we noticed that $\mu$ does not
contain enough information to determine $*$.
The mapping 
\mbox{$\nu \eqdef \varphi \longmapsto \rev{\rev{K}{\!*}{\psi}}{*}{\varphi}$} 
is not determined by \rev{K}{\!*}{\psi} and $\mu$ alone.
The method proposed in this paper for revising with a rational,
consistency-preserving relation does not yield any method for adapting
revisions, any {\em transmutation} in the terms of~\cite{MaryAnneW:trans}. 
\subsection{Conclusion and future work}
\label{subsec:conc}
Building on previous work by many researchers, we have completed the
analysis of the relation between the extended AGM postulates
for theory revision and properties of nonmonotonic inference.
It was known that one may associate rational, consistency-preserving relations
to any revision that satisfies the AGM postulates.
We have shown that any such relation is associated with exactly one revision
that satisfies the AGM postulates and an additional one.
The open questions on which further work is needed have to do with
generalizing the results of this paper to larger classes of revisions
and nonmonotonic systems.

One may ask if our results can be generalized to the revisions that
satisfy only the postulates: \bKs{1}--\bKs{7}.
Those revisions are known to be intimately related to preferential
relations~\cite{Rott:92}. 
Preferential relations may stand in one-to-one correspondence with 
the revisions that satisfy \bKs{1}--\bKs{7} and \bKs{9}.

An orthogonal avenue seems also interesting. 
We noticed that \bKs{3} and \bKs{4} have no influence on the nonmonotonic
aspect of revision. 
The revisions that satisfy \bKs{1}, \bKs{2} and \bKs{5}--\bKs{8}
seem an interesting class to study.

We noticed that, in the AGM framework, the theories to be revised
were unstructured sets of formulas. One would like to put some more
structure on those theories. 
Such additional structure, on $K$, should probably constrain the revisions
of the revised theories of the form \rev{K}{\!*}{\varphi}.
The analogy between (unstructured) theory revision and nonmonotonic
inference should probably lift to an analogy between generalized theory
revision and conditional logic. 
\section{Acknowledgment}
During the last stage of the elaboration of this work,
Daniel Lehmann held long and fruitful conversations with Isaac Levi.
They are gratefully acknowledged.
David Makinson's detailed and knowledgeable comments on a first draft
helped us catch up on the history of the field and improve the presentation
altogether.

\end{document}